\documentclass{dukecei}

\usepackage{enumitem}
\usepackage{threeparttable}
\usepackage{hyperref}
\usepackage{makecell}
\usepackage{colortbl,xcolor}
\usepackage{longtable}
\usepackage{listings}
\usepackage{tcolorbox}

\newcommand{\Fig}[1]{Fig.~\ref{#1}}

\newcommand{\Tbl}[1]{Tbl.~\ref{#1}}
\newcommand{\Sec}[1]{Sec.~\ref{#1}}

\newcommand{\Reb}[1]{\textcolor{red}}

\newcommand{\pad}{\textit{DPad}}

\definecolor{vscodedef}{RGB}{0,0,0}         
\definecolor{vscodekw}{RGB}{0,0,205}        
\definecolor{vscodecomment}{RGB}{0,128,0}   
\definecolor{vscodestring}{RGB}{163,21,21}  
\definecolor{vscodeframe}{RGB}{200,200,200} 

\title{DPad: Efficient Diffusion Language Models with Suffix Dropout}

\author{Xinhua Chen$^{1}$\footnote{Equal contribution},
    Sitao Huang$^{1}$\footnotemark[1],
    Cong Guo$^{1}$\footnotemark[1]\footnote{Corresponding author: Cong Guo (cong.guo@duke.edu)},
    Chiyue Wei$^{1}$,
    Yintao He$^{1}$,
    Jianyi Zhang$^{1}$,
    Hai~``Helen''~Li$^{1}$,
    Yiran Chen$^{1}$    
    \\[1em]
    \normalsize $^{1}$Duke University
}

\begin{document}

\maketitle
\thispagestyle{firstpagestyle} 

\begin{abstract}
Diffusion-based Large Language Models (dLLMs) parallelize text generation by framing decoding as a denoising process, but suffer from high computational overhead since they predict all future suffix tokens at each step while retaining only a small fraction.  
We propose \textbf{Diffusion Scratchpad} (\textbf{\pad{}}), a training-free method that restricts attention to a small set of nearby suffix tokens, preserving fidelity while eliminating redundancy.  
\pad{} integrates two strategies: (i) a \textit{sliding window}, which maintains a fixed-length suffix window, and (ii) \textit{distance-decay dropout}, which deterministically removes distant suffix tokens before attention computation.  
This simple design is compatible with existing optimizations such as prefix caching and can be implemented with only a few lines of code.  
Comprehensive evaluations across multiple benchmarks on \texttt{LLaDA-1.5} and \texttt{Dream} models demonstrate that \pad{} delivers up to $\mathbf{61.4\times}$ speedup over vanilla dLLMs while maintaining comparable accuracy, highlighting its potential for efficient and scalable long-sequence inference.  
Our code is available at \url{https://github.com/Crys-Chen/DPad}.
\end{abstract}

\section{Introduction}
\begin{figure}[t]
    \centering
    \includegraphics[width=0.98\linewidth]{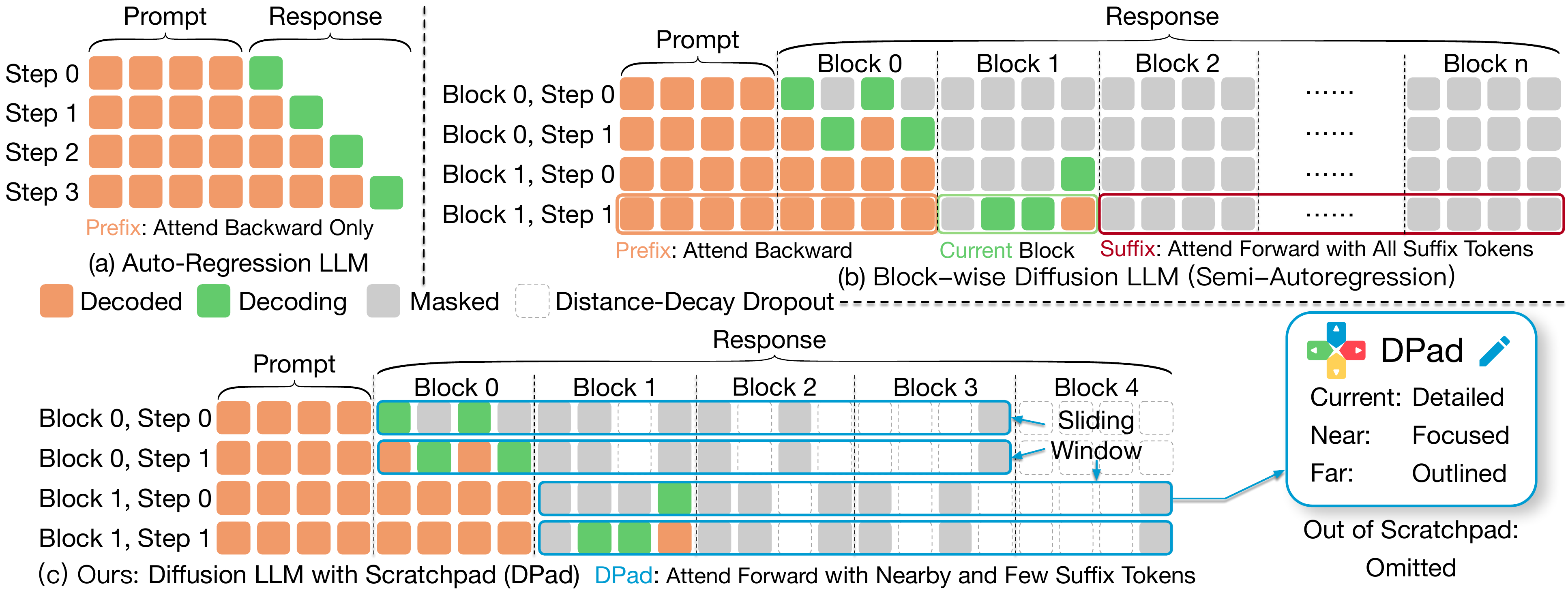}
    \caption{
    Comparison of (a) autoregressive LLMs, (b) block-wise diffusion LLMs, and (c) our \pad{}.  
    \pad{} restricts suffix attention via:  
    (\textit{i}){Sliding Window}: fixed-length suffix window;  
    (\textit{ii}) {Distance-decay Dropout}: removes distant suffix tokens without computing attention scores.
    }
    \label{fig:intro}
\end{figure}

Large Language Models (LLMs) have become foundational in numerous applications~\cite{vaswani2017attention,devlin2019bertpretrainingdeepbidirectional,brown2020languagemodelsfewshotlearners,ouyangTraining2022}, yet their deployment is often hindered by inference latency. 
As shown in \Fig{fig:intro}~(a), the predominant autoregressive framework generates text one token at a time~\cite{radford2018language,radford2019language}, imposing a sequential constraint that limits speed and scalability~\cite{gu2018nonautoregressive}.  
This has driven interest toward parallel decoding strategies.

Diffusion-based Large Language Models~\cite{li2022diffusionlm,austin2021structured,lou2024discrete,shi2025simplifiedgeneralizedmaskeddiffusion,israel2025acceleratingdiffusionllmsadaptive} (dLLMs) offer a promising alternative by eliminating sequential dependencies.
Formulating text generation as a parallel denoising process, they can predict entire sequences or generate text block-wise (i.e., semi-autoregressively)~\cite{llada,dream2025}, as in \Fig{fig:intro}~(b).  
However, this parallelism often incurs high computational cost~\cite{nie2025scalingmaskeddiffusionmodels}: at each step, predictions for \textbf{\emph{all}} future (suffix) tokens are computed, though only a small fraction are retained.  
Consequently, although dLLMs can generate multiple tokens in parallel, the resulting throughput gains are undermined by a disproportionate increase in computation, posing a key bottleneck to their widespread adoption~\cite{song2025seed-full}.

To further understand this inefficiency, we analyze the role of suffix tokens under the block-wise masking mechanism in dLLMs. Interestingly, these suffix tokens contribute no direct semantic information to the attention; instead, they function as an \textit{information reservoir} that collects signals from the already decoded prefix tokens, much like a scratchpad. 
Through multiple Transformer layers, this scratchpad can guide the diffusion process to produce a more organized and contextually consistent current block.  
Yet, most suffix tokens are redundant and low-entropy, often forming long runs of similar content.  
This redundancy becomes more severe as the distance to the current block increases, with attention scores for suffix tokens dropping sharply from near to far positions. 
Such redundancy not only imposes unnecessary computational overhead, but can also degrade the fidelity of the generated content.

Based on the above insights, we propose the \textbf{Diffusion Scratchpad} (\textbf{\pad{}}), which attends forward only to a small number of near suffix tokens, as in \Fig{fig:intro}~(c).  
It uses two suffix drop strategies: \textit{sliding window} and \textit{distance-decay dropout}.  
For the \textbf{\textit{sliding window}}, inspired by prefix KV-cache optimizations~\cite{beltagy2020longformer, xiao2023efficient}, we extend the idea to the suffix.  
Here, the suffix window has a fixed length and moves forward along with the current block, retaining only a limited number of nearby suffix tokens.  
In contrast to vanilla dLLMs, where suffix computation increases with the generation sequence length, our design keeps the cost bounded and reduces suffix-related computation by one dimension.

For the \textbf{\textit{distance-decay dropout}}, suffix tokens are removed according to their distance from the current block: the farther they are, the higher the dropout ratio, until all tokens beyond the window are omitted.  
Unlike conventional attention-score pruning~\cite{Wang_2021,kim2022learnedtokenpruningtransformers,song2025sparsedllmacceleratingdiffusionllms}, which first computes attention scores and then prunes based on their magnitude, \pad{} predetermines a distance-decay sparse pattern for suffix tokens \emph{prior to} model execution and eliminates them in a single operation at the beginning of each step.  

Inspired by the Lottery Ticket Hypothesis (LTH)~\cite{frankle2018the}, which posits that properly pruned sub-networks with their original initialization can match the performance of dense networks after training, we observe that the strong generalization ability of dLLMs enables the \emph{training-free} construction of ``winning tickets'' for suffix tokens during inference. 
Although this predetermined dropout strategy may appear intrusive, it consistently achieves strong empirical performance, with only a small subset of suffix tokens required to maintain accuracy comparable to vanilla dLLMs.  
This further suggests that sparsity is an inherent property of suffix attention.  
Additionally, the method is extremely simple to deploy, requiring only a few lines of code to implement.

To illustrate the intuition behind \pad{}, we liken it to a real scratchpad used when writing a book, as shown in \Fig{fig:intro}~(c, right).  
For the current chapter (i.e., block), the writer (dLLM) devotes the most attention, revising it multiple times, akin to denoising within a diffusion block.  
The upcoming chapter receives focused drafts for consistency, while much later chapters contain only brief outlines.  
Naturally, the ``writer'' should not, and indeed must not,  fill all future chapters (all suffix tokens) with low-entropy, repetitive content merely to satisfy the fixed sequence length constraints of current dLLMs.  
Such uncontrolled filling distracts the ``writer's'' attention and wastes storage and computation, making it neither sustainable nor scalable.

Finally, we emphasize that \pad{} enables efficient and compact generation without enforcing a fixed sequence length.  
Unlike conventional dLLMs, where mandatory sequence length and full suffix attention introduce redundancy and degrade output quality~\cite{dream2025,llada}, \pad{} produces concise results with minimal overhead.  
It is entirely \textit{training-free}, saving substantial memory and computation, and achieves stable acceleration even at shorter generation sequence lengths (e.g., $256$ and $512$), outperforming both vanilla and optimized variants such as Fast-dLLM~\cite{fast-dllm}.  
\pad{} is also compatible with existing optimizations~\cite{fast-dllm,dllm-cache,slow_fast_sampling}, including prefix caching~\cite{fast-dllm}, dual caching~\cite{fast-dllm}, and dLLM-Cache~\cite{dllm-cache}.  
It further scales naturally to long sequences: when combined with parallel generation and prefix caching, it achieves a $\mathbf{30.58\times}$ speedup on \texttt{Dream}/\texttt{HumanEval} ($1024$, 0-shot) and a $\mathbf{61.39\times}$ speedup on \texttt{LLaDA-1.5}/\texttt{GSM8K} ($1024$, 1-shot) over vanilla, while maintaining comparable model accuracy. 
These results highlight the strong potential of \pad{} for long-sequence inference and demonstrate its ability to unlock new efficiency frontiers in dLLMs.  
Although dLLMs remain in an early exploratory stage, with many researchers striving to assemble the final puzzle, we believe that \pad{} represents one of the key components in completing the whole picture and paves the way toward practical, scalable diffusion-based language modeling.

\section{Preliminary}

\subsection{Foundational Principles of Diffusion Large Language Models (dLLM)}
\label{dLLMfoundamentals}

Conventional Autoregressive Models (ARMs) define the joint probability of a generated sequence $x = (x_1, \dots, x_L)$ by factoring it into a product of conditional probabilities in a fixed sequential order~\cite{vaswani2017attention}. The model predicts the next token $x_i$ based on the conditional probability of the preceding tokens $x_1, \dots, x_{i-1}$:
\begin{equation}
    p_\theta(x) = p_\theta(x_1) \prod_{i=2}^{L} p_\theta(x_i | x_1, \dots, x_{i-1})
\end{equation}

In contrast, Diffusion Language Models (dLLMs) leverage a non-autoregressive process ~\cite{austin2021structured,shi2025simplifiedgeneralizedmaskeddiffusion,lou2024discrete}. During training, the model learns to denoise a corrupted sample. This involves two processes:

\paragraph{Forward Masking Process}This process systematically replaces a proportion of tokens in a clean text sequence $x_0$ with a special \texttt{[MASK]} token ~\cite{nie2025scalingmaskeddiffusionmodels,llada}, similar to applying noise in a conventional diffusion model ~\cite{ho2020denoisingdiffusionprobabilisticmodels}. The replacement is governed by a masking schedule where $t \in [0, 1]$ represents the masked level. For a given $t$, the probability of an individual token $x_0^i$ being masked is:
\begin{equation}
q(x_t^i | x_0^i) = \begin{cases} 1-t, & \text{if } x_t^i = x_0^i \\ t, & \text{if } x_t^i = \text{\texttt{[MASK]}} \end{cases}
\end{equation}

\paragraph{Reverse Unmasking Process} The model $p_\theta$ is trained to predict the original tokens given the partially masked sequence $x_t$, thereby learning to approximate the true unmasking probability $q(x_0 | x_t)$ ~\cite{austin2021structured,lou2024discrete}. This is achieved by minimizing the negative log-likelihood over the masked tokens~\cite{shi2025simplifiedgeneralizedmaskeddiffusion,ouyangTraining2022}:
\begin{equation}
\mathcal{L}(\theta) = -\mathbb{E}_{t, x_0, x_t} \left[ \sum_{i: x_t^i = \text{\texttt{[MASK]}}} \log p_\theta(x_0^i | x_t) \right]
\end{equation}
This process makes use of a Transformer architecture to learn the distribution $p_\theta(x_0 | x_t)$. During training, bidirectional multi-head attention on all tokens within a sequence---masked and unmasked tokens alike---is employed to predict the probability distribution for each masked token.

\subsection{Inference and Sampling in dLLM}

The inference process begins by initializing a sequence $y_0$ through the concatenation of a prompt $r$ and $L$ mask tokens, denoted as $y_0 = r \circ (\texttt{[MASK]})_{i=1}^L$. Let $M_s$ be the set of indices corresponding to masked tokens at step $s$; initially, $M_0 = \{|r|+1, \dots, |r|+L\}$. The model then iteratively refines this sequence over steps $s=1, 2, \dots, S$. At each step $s$, the model $p_\theta$ computes a probability distribution $p_\theta(y_i | y_{s-1})$ for every masked position $i \in M_{s-1}$. From these distributions, the most likely token predictions $\hat{y}_i$ and their corresponding confidence scores $c_i$ are determined:

\begin{equation}
    \hat{y}_i = \underset{v \in V}{\text{argmax}} \, p_\theta(y_i=v | y_{s-1}) \quad \text{and} \quad c_i = p_\theta(y_i=\hat{y}_i | y_{s-1}).
\end{equation}
where $V$ is model's vocabulary.
A scheduling function $\mathcal{G}(s, S)$ determines the number of tokens, $k_s$, to unmask. The set of indices to update, $U_s$, is chosen by selecting the $k_s$ positions from $M_{s-1}$ with the highest confidence scores. The new sequence $y_s$ is then formed by updating these positions:
\begin{equation}
(y_s)_i = \begin{cases} \hat{y}_i & \text{if } i \in U_s \\ (y_{s-1})_i & \text{otherwise} \end{cases}
\end{equation}
The set of masked indices is updated via $M_s = M_{s-1} \setminus U_s$. This iterative refinement continues until $M_s = \emptyset$.

\subsection{Related work on dLLM acceleration}

While early work in dLLMs employed Top-$k$ sampling to unmask a fixed number of high-confidence tokens at each step, later methods introduced more dynamic, confidence-aware techniques~\cite{fast-dllm,slow_fast_sampling} that greedily unmask all tokens above a threshold, thereby alleviating the generation bottleneck.  
Another line of research focuses on cache management.  
Because attention in dLLMs is bi-directional, conventional KV caching is not directly applicable.  
To address this, several optimizations~\cite{fast-dllm,dllm-cache} exploit the observation that prefix tokens and distant suffix tokens exhibit little variation in their attention values across inference steps, allowing them to be reused and reducing expensive recomputation.  
A more fine-grained approach~\cite{song2025sparsedllmacceleratingdiffusionllms} further improves efficiency by dynamically evicting cache entries based on attention scores, leveraging the persistence of attention patterns across steps.

\section{Method}
In this section, we begin by formalizing the {\emph{Scratchpad}} mechanism in dLLMs, which underpins the role of suffix tokens as an information reservoir.  
We then introduce {\emph{Suffix Dropout}}, which combines a fixed-length sliding window that moves with the current block to bound the number of attended suffix tokens, with a distance-decay scheme that progressively prunes more distant ones.  
Finally, we propose the {\emph{Diffusion Lottery Tickets}} (DLT) hypothesis, which provides a principled lens for interpreting dropout in dLLMs.

\begin{figure}[t]
    \centering
    \includegraphics[width=0.95\linewidth]{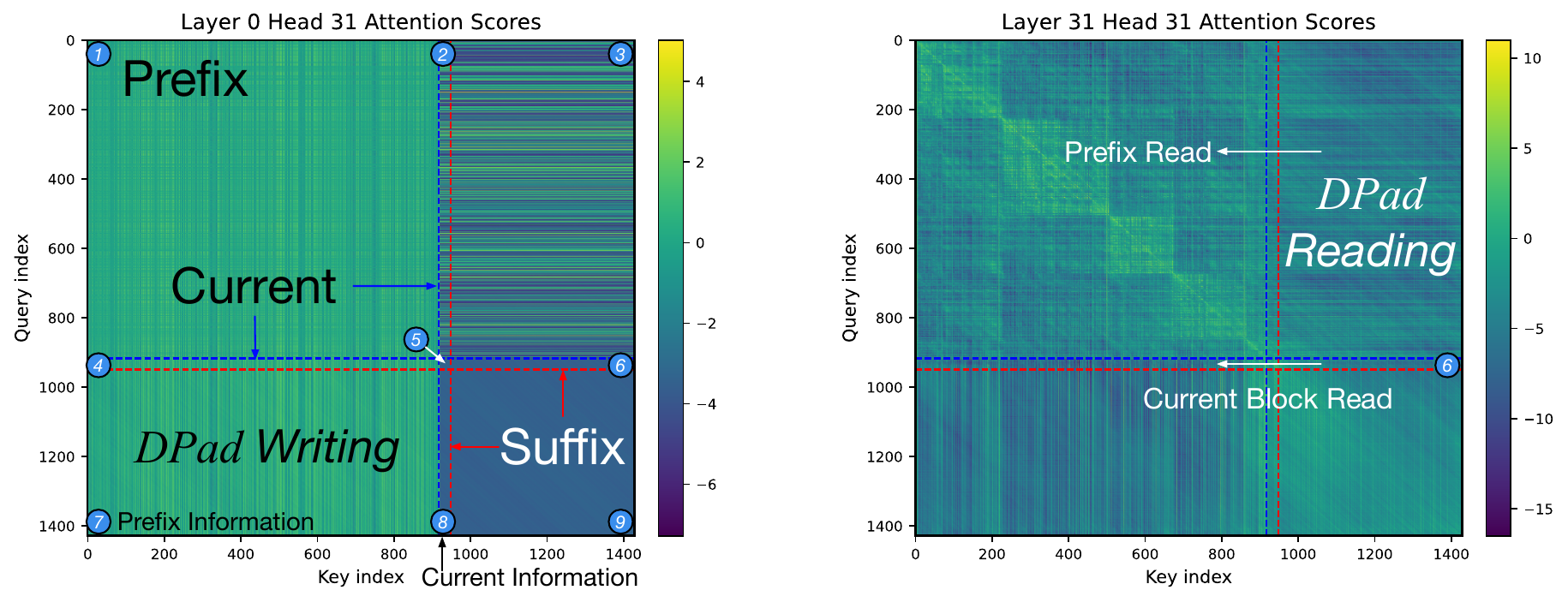}
    \caption{
    Attention score maps illustrating the \textbf{Scratchpad} mechanism in dLLMs. The maps were generated by the LLaDA-1.5 model \cite{zhu2025llada15variancereducedpreference} on prompt and 512-token sequences from the GSM8K dataset \cite{cobbe2021gsm8k}.
    The attention matrix is divided into $3\times 3$ blocks over \textit{prefix}, \textit{current}, and \textit{suffix}.  
    Blocks 7 and 8 collect information from the prefix and current into the suffix at layer $n$, while Block 6 feeds this stored information back to the current block at layer $(n\!+\!1)$.  
    This write–store–read cycle makes suffix tokens a dynamic scratchpad for cross-layer contextual reuse.
    }
    \label{fig:dpad_attention}
\end{figure}

\subsection{Scratchpad Mechanism}
\label{scratchpad}
We first revisit the role of suffix tokens in dLLMs.  
Due to the masking structure, suffix tokens carry no direct semantic information but instead serve as an \emph{information reservoir}, aggregating signals propagated from prefix tokens across multiple Transformer layers.  
This latent memory, which we refer to as \textbf{\pad{}}, stabilizes the denoising process by providing contextual scaffolding for the current block.  

As illustrated in Fig.~\ref{fig:dpad_attention} and the attention score maps, the token sequence can be partitioned into three contiguous segments: Prefix indices $[0, c-1]$, current indices $[c, s-1]$, and suffix indices $[s, L-1]$.  
The corresponding attention matrix is thus divided into $3 \times 3$ blocks.  
Among them, Blocks 6, 7, and 8 together define the scratchpad mechanism.  

Considering only one head, at layer $n$, queries from prefix and current tokens attend to keys from the suffix region.  
Formally, the global attention weights are defined as
\begin{equation}
A^{(n)} = \text{Softmax}\!\left(\frac{Q^{(n)} (K^{(n)})^\top}{\sqrt{d_k}}\right) \in \mathbb{R}^{L \times L}.
\end{equation}
We can partition $A^{(n)}$ into submatrices corresponding to prefix ($P = [0,c-1]$), current ($C = [c,s-1]$), and suffix ($S = [s,L-1]$).  
In particular,
\begin{equation}
A^{(n)}_{S,P} = A^{(n)}[s:L,\, 0:c],
\qquad
A^{(n)}_{S,C} = A^{(n)}[s:L,\, c:s],
\end{equation}
represent the attention weights from suffix queries to prefix and current keys, respectively (Blocks 7 and 8 in Fig.~\ref{fig:dpad_attention}).  
Multiplying these attention weights with the value matrix yields the actual outputs at suffix positions:
\begin{equation}
H^{(n)}_{S} = A^{(n)}_{S,P} V^{(n)}_{P} \;+\; A^{(n)}_{S,C} V^{(n)}_{C} \;+\; A^{(n)}_{S,S} V^{(n)}_{S}.
\end{equation}

Here, $H^{(n)}_{S}$ denotes the hidden representations of suffix tokens after attention.  
This equation shows that suffix tokens integrate information from prefix and current regions, effectively serving as a \emph{Scratchpad} that records contextual signals.  
After this aggregation, the outputs $H^{(n)}$ are processed by the subsequent linear transformations (e.g., feed-forward layers and residual connections), which operate independently on each token.  
Thus, information written into the suffix tokens will not interact with other regions until the next layer’s attention, where prefix and current queries can \emph{read} back the stored information from \pad{}.  
This establishes the complete write–store–read cycle across consecutive layers.
At layer $(n\!+\!1)$, this stored information can be retrieved by the current block through
\begin{equation}
A^{(n+1)}_{C,S} = A^{(n+1)}[c:s,\, s:L],
\end{equation}
which corresponds to current-to-suffix attention (Block 6 in Fig.~\ref{fig:dpad_attention}).  
This path enables the current block to reuse the information collected in the suffix at the previous layer. 

Through this mechanism, the stored information in suffix tokens at layer $n$ can be retrieved by the prefix and current blocks at layer $(n\!+\!1)$.  
However, in practice, the influence of the suffix on the prefix is negligible: Prefix tokens correspond to already generated content, and their representations can be refined directly from themselves and the current block, without requiring additional signals from the suffix.  
Indeed, prior work~\cite{fast-dllm} has shown that caching prefix states during inference barely impacts accuracy, implying that the suffix-to-prefix pathway is effectively redundant.  
Therefore, the essential interaction of the scratchpad mechanism lies in the current-to-suffix direction (Block 6), where the suffix serves as temporary memory to assist the ongoing denoising process.

Therefore, we conjecture that the behavior of suffix tokens resembles a \textbf{\emph{residual connection}} specialized for attention.  
Instead of directly propagating representations, the model compresses high-dimensional signals from the prefix and current into the suffix and re-injects them into the current block at the next layer.  
This is reminiscent of the \emph{bottleneck residual block} in ResNets~\cite{he2015deepresiduallearningimage}, where high-dimensional features are projected, propagated, and then expanded.  
Similarly, the suffix acts as a scratchpad that compresses contextual information and delivers it forward for cross-layer reuse in dLLMs.

\subsection{Suffix Dropout Strategies}
\begin{figure}[t]
    \centering
    \includegraphics[width=0.95\linewidth]{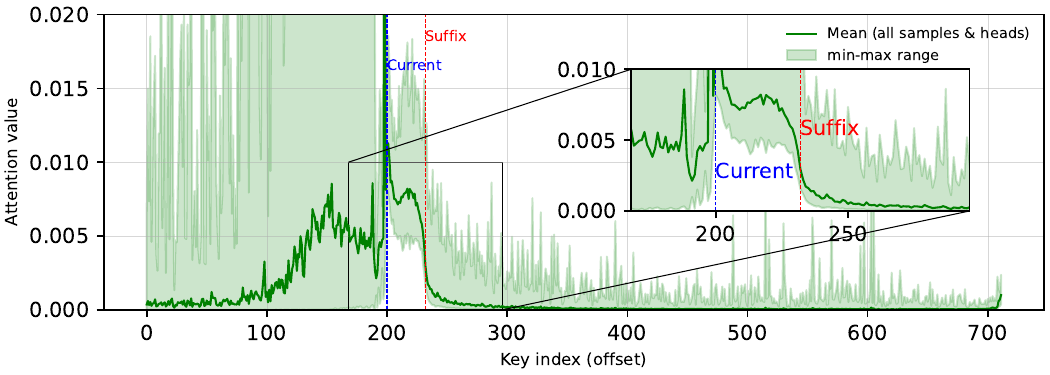}
    \caption{
    Analysis of the {Suffix Drop strategy} at the final layer (Layer 31), generated by the LLaDA-1.5 model \cite{zhu2025llada15variancereducedpreference} on GSM8K dataset~\cite{cobbe2021gsm8k} with a max length of 512.
    We collect attention weights from 100 samples across all heads, focusing on the current block queries ($A$[$c$:$s$, $c$-200:]).  
    To align positions, we truncate $200$ tokens before $c$ as the prefix and align all current blocks at $c$.  
    The plot shows the mean attention distribution over key indices (green curve), together with the min–max range (shaded area).  
    Current and suffix boundaries are marked, showing that attention on far suffix tokens rapidly decays, motivating our distance-decay dropout design.
    }
    \label{fig:suffix_drop_scores}
\end{figure}

Building on the scratchpad perspective, we analyze redundancy in suffix attention to motivate a structured dropout design.  
At the final layer 31, \Fig{fig:suffix_drop_scores} shows a sharp distance-dependent decay: nearby suffix tokens dominate, consistent with general attention behavior and prior work on prefix sliding windows~\cite{beltagy2020longformer,xiao2023efficient}.  
Moreover, suffix attention is one to two orders of magnitude smaller than that of prefix or current, indicating that suffix representations are information-light and only a small subset is needed to serve as the scratchpad memory.  

Based on these findings, we propose two complementary suffix drop strategies:  
(a) fixed-length \emph{Sliding Window} to retain only a bounded number of near-suffix tokens,  
and (b) \emph{Distance-decay Dropout} that progressively prunes distant suffix tokens with minimal impact on model quality,  
as illustrated in \Fig{fig:intro}(c).  
Both mechanisms are efficiently realized through a Gaussian sampling process, which simultaneously enforces a bounded window and distance-dependent decay.  

As a result, suffix attention only needs to focus on nearby tokens, making the suffix dropout window decoupled from the overall sequence length.  
Unlike the vanilla setting, where the suffix grows with the generation sequence, our approach keeps it constant.  
This yields a clear computational benefit: suffix dropout effectively reduces suffix-related complexity by one dimension.  
Formally, let $L$ denote the suffix length and $B$ the block size.  
In vanilla block-wise dLLMs, suffix computation requires $\tfrac{L}{B}$ steps, each involving $O(L)$ suffix token operations, whereas suffix dropout restricts this to a constant number of tokens, independent of $L$.

\paragraph{Gaussian Sampling.}
Inspired by the smooth decay patterns observed in \Fig{fig:suffix_drop_scores}, we adopt a Gaussian sampling strategy to implement distance-aware token dropout.  
Formally, for a suffix token at distance $d$ from the suffix boundary, its retention probability $P(d)$ is defined by the right half of a standard normal distribution with effective window size $W$:  
\begin{equation}
\label{eq:gaussian_dropout}
P(d) = a \cdot \frac{1}{\sigma \sqrt{2\pi}} 
\exp\!\left[-\tfrac{1}{2}\left(\tfrac{\tfrac{k\sigma}{W}\cdot d -\mu}{\sigma}\right)^2\right], 
\quad 0 < d \leq W,
\end{equation}
where $\mu=0$, $\sigma=1$, and the suffix boundary is the center of the Gaussian distribution.  
Two hyperparameters control the distribution along the $x$- and $y$-axes:  
(1) $\tfrac{k\sigma}{W}$ maps the window size $W$ to $k\sigma$ (e.g., for $W=256$, setting $k=3$ ensures $d=256$ corresponds to $3\sigma$);  
(2) $a$ scales the overall sampling magnitude vertically.  
This formulation ensures that tokens farther from the boundary are retained with exponentially decreasing probability, implemented via rejection sampling.

This dropout strategy is not static but is dynamically re-applied in a block-wise way. As shown in \Fig{fig:intro}~(c), after a block is generated, all previously dropped-out tokens are restored. The Gaussian sampling is then invoked again on the updated set of suffix tokens before the next block is decoded. This dynamic resampling is crucial for preventing sampling bias and ensuring that no token is systematically ignored during the generation process.

\paragraph{Implementation.}  
Our method can be implemented in just a few lines of code: \begin{tcolorbox}[colback=black!0,boxrule=0pt,arc=2pt,
  left=0pt,right=0pt,top=0pt,bottom=0pt]
\begin{lstlisting}[language=Python]
# Select suffix token indices with Gaussian distance-decay sampling
keep_idx = gaussian_suffix_dropout(x)
# Preserve only the selected tokens
x_pruned = x[keep_idx]
# Forward the reduced sequence to the model
output = model(x_pruned, args)
\end{lstlisting}
\end{tcolorbox}
Here, the Gaussian-based sampler \texttt{gaussian\_suffix\_dropout} selects a subset of suffix tokens according to distance-aware probabilities.  
We then construct a pruned sequence \texttt{x\_pruned} by indexing with \texttt{keep\_idx}.  
The rest of the model remains unchanged, except for a minor adjustment to the RoPE embeddings inside the attention module to ensure correct positional encoding under suffix dropout.

Besides, we also introduce an early termination mechanism to prevent inefficient computation after a \texttt{<eos>} token is generated, a common issue in fixed generation length strategies like LLaDA~\cite{llada}. Our method performs a conditional check after each decoded block; if an \texttt{<eos>} token is found, we fill the remaining sequence with \texttt{<eos>} and immediately halt generation.

\paragraph{RoPE Adjustment.}Our suffix dropout mechanism requires only a minor adjustment to the Rotary Positional Embedding (RoPE)~\cite{su2021roformer} to maintain correct positional information.  
In standard RoPE, a token at absolute position $i$ is encoded using an angle $\theta_i = i \cdot \Delta$.  
After dropout, however, only a sparse, non-contiguous subset of suffix indices $\mathcal{I} = \{i_1, \dots, i_m\}$ is preserved.  

To handle this, we ensure that each preserved token retains its original positional information.  
Rather than using their re-indexed positions after dropout, we apply a mapping function $f(i_k)$ that retrieves the original absolute position of the $k$-th preserved token.  
The new angle is then computed as
\begin{equation}
\theta'_{i_k} = f(i_k) \cdot \Delta .
\end{equation}
Accordingly, the modified RoPE application becomes
\begin{equation}
\text{RoPE}'(\mathbf{x}_{i_k}, i_k) = \text{RoPE}(\mathbf{x}_{i_k}, f(i_k)).
\end{equation}
This adjustment requires only a lightweight remapping inside the attention module and does not alter the functional form of RoPE, confirming that suffix dropout is nearly cost-free while preserving positional consistency.

\begin{figure}
\centering
\includegraphics[width=1\textwidth]{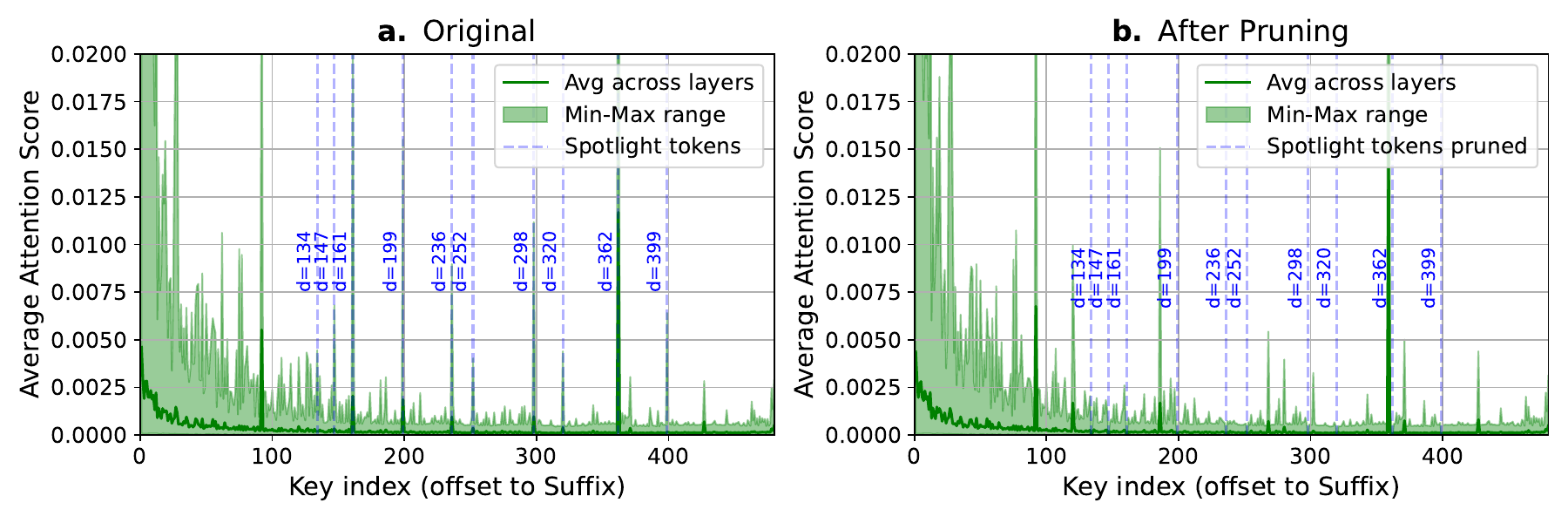}
\caption{(a) Average, maximum, and minimum attention scores of suffix tokens paid by current block tokens ($A$[$c$:$s$, $s$:]) across layers in LLaDA-1.5, showing overall decay with occasional spikes (e.g., d = 199, 298, 362). (b) After forcibly pruning these spike positions, attention shifts to nearby tokens, indicating that adjacent positions can absorb suffix information (e.g., pruning token 362 shifts the spike to token 359).
}
\label{fig:attention_score}
\end{figure}

\subsection{Diffusion Lottery Tickets Hypothesis}
The analysis in Fig.~\ref{fig:suffix_drop_scores} not only reveals the overall decay of current-to-suffix attention, but also occasional sharp \emph{spikes} in the maximum values.  
These spikes cannot be predicted in advance and, in principle, pruning them could affect model accuracy.  
To investigate this, we conduct an additional experiment shown in Fig.~\ref{fig:attention_score}.  
We first run dLLM inference for one step, then forcibly prune the top 10 highest-attention suffix tokens (``spotlight'' tokens) occurring beyond the first 128 positions (e.g., at distances 199, 298, and 362).  
Surprisingly, pruning such distant suffix tokens, even those corresponding to large spikes, has little effect on model accuracy.  
When these spikes are removed, the model shifts its attention to nearby suffix tokens, for example, the spotlight token at \textbf{362} in Fig.~\ref{fig:attention_score}~(a) is replaced by increased attention at its neighbor token \textbf{359} in Fig.~\ref{fig:attention_score}~(b), which effectively \emph{absorbs} the lost information.  
This behavior is consistent with the strong generalization ability of diffusion models and the fact that suffix tokens are initialized without semantic content: through \pad{} mechanism, suffix tokens can dynamically learn and store information during inference.  
Consequently, information carried by distant suffix tokens appears largely \emph{position-insensitive}, and pruning spike positions has almost no impact on final accuracy, as further confirmed in our evaluation experiments.  

This leads to an important intuition: information collected by distant suffix tokens is insensitive to their exact positions.  
When writing a book, one may use a scratchpad to record an event in the “middle chapters,” but it is not necessary to pin it down to precisely token 186; recording it at token 152 or 198 suffices.  
This explains why suffix dropout can be applied \emph{a priori}, without computing exact attention scores, and why it fundamentally differs from prefix cache pruning~\cite{song2025sparsedllmacceleratingdiffusionllms,Wang_2021}:  
Prefix tokens carry dense, position-bound semantic information and thus cannot be arbitrarily discarded, whereas suffix tokens act as a flexible, low-rank memory buffer.

These observations resonate with the \emph{Lottery Ticket Hypothesis (LTH)}~\cite{frankle2018the}, which posits that properly pruned sub-networks with their original initialization can match the performance of dense networks after training.  
We extend this intuition to dLLMs and propose the \textbf{Diffusion Lottery Tickets (DLT)} hypothesis:  
During inference, the suffix region contains redundant tokens, yet a sparse subset is sufficient to preserve semantic consistency and generation quality.  
Through \pad{} mechanism, this subset can be adaptively reorganized into ``winning tickets'' within the forward pass.  
In this view, suffix dropout becomes a \emph{training-free lottery ticket search}, where Gaussian sampling selects a compact set of suffix tokens that carry the essential information for denoising in dLLMs.

\paragraph{Summary.}  
We showed that suffix tokens function as a scratchpad, enabling a high-dimensional residual pathway.  
This insight motivates two lightweight strategies, {Sliding Window} and {Distance-decay Dropout}, together with the {DLT hypothesis}, which interprets suffix pruning as a training-free lottery ticket search.  
These methods are simple, training-free, and easy to integrate.  
Next, we evaluate the efficiency and accuracy of \pad{}-enhanced dLLMs across standard benchmarks.

\section{Experiments}
\subsection{Experimental Setup}

\paragraph{Models and Baselines.} 
All experiments are conducted on an NVIDIA A100 80GB GPU.
We evaluate \pad{} on a suite of representative open-source dLLMs: two variants of LLaDA (8B-Instruct and 1.5)~\cite{llada,zhu2025llada15variancereducedpreference} and Dream-v0-Base-7B~\cite{dream2025}.  
We compare against the following baselines:

\begin{itemize}[nosep]
    \item \textbf{Vanilla}: the unmodified LLaDA~\cite{llada} and Dream~\cite{dream2025} backbones, which compute attention over all tokens in the sequence, including prefix, current, and suffix tokens.  
    \item \textbf{+Parallel (Fast-dLLM)}: a parallel decoding strategy applied to the vanilla backbone, where tokens are unmasked whenever their confidence exceeds a predefined threshold~\cite{fast-dllm}, instead of the fixed top-$k$ unmasking used in Vanilla.  
    \item \textbf{+Prefix Cache}: applying KV caching on prefix tokens, leveraging the observation that their key–value states remain nearly static across inference steps~\cite{fast-dllm}, thereby avoiding redundant computations.  
\end{itemize}

When combining \pad{} with caching, we adopt a standard Prefix Cache.  
Since our dropout already minimizes the computational cost of the suffix, the Dual Cache mechanism proposed in~\cite{fast-dllm} is unnecessary.  

The \pad{} strategy introduces three hyperparameters: a decay rate factor $k$, a magnitude scalar $a$, and a sliding window size.  
We tuned these on small subsets of each benchmark (see \Sec{sec:window_function}).  
Unless otherwise specified, we use a block size of 32 and set the confidence threshold for parallel decoding to 0.9.    

\paragraph{Benchmarks and Metrics.}  
Our evaluation spans multiple domains.  
For reasoning, we use GSM8K~\cite{cobbe2021gsm8k} and MATH~\cite{hendrycksmath2021};  
for code generation, we use HumanEval~\cite{humaneval} and MBPP~\cite{mbpp}.  
We report two aspects of performance:
We report two aspects of performance:
\begin{itemize}[nosep]
    \item \textbf{Accuracy}: evaluated with task-specific metrics, including pass@1 for code generation and accuracy measures for reasoning (e.g., flexible-extract and strict-match accuracy on GSM8K). All accuracies are reported as percentages.
    \item \textbf{Efficiency}: measured by average inference latency per sample and Tokens Per Second (TPS). TPS is computed over the entire generated sequence until an <eos> token is produced. 
    We also use $a$ / $b$ to denote the generation (Gen.) length, where $a$ is the average real generation length and $b$ is the benchmark setting length.
\end{itemize}

\subsection{Main Results} 
\label{sec:main_results}
We evaluate the efficiency and accuracy of \pad{} on three models and four benchmarks, as reported in \Tbl{tb:instruct}, \Tbl{tb:1.5}, and \Tbl{tb:dream}.  
We compare against both the vanilla LLaDA baseline~\cite{llada} and the parallel decoding baseline Fast-dLLM~\cite{fast-dllm}.  
Our results reveal three consistent trends:
\begin{itemize}[nosep]
    \item \textbf{Latency:} consistently improved across all settings. 
    \item \textbf{Throughput (TPS):} shows higher variance, with fluctuations across benchmarks and models.   
    \item \textbf{Accuracy:} Flexible Match remains comparable, while Strict Match achieves significant gains.  
\end{itemize}
We analyze each of these aspects in detail below.

\begin{table}[t]
\footnotesize 
\setlength{\tabcolsep}{4pt} 
\centering
\caption{Performance of LLaDA-Instruct with \pad{} on four benchmarks.}
\label{tb:instruct}
\begin{tabular}{@{}l l rrrr r rr@{}}
\toprule
\multicolumn{9}{c}{LLaDA-Instruct
} \\
\midrule
\multirow{2}{*}{Benchmark} & \multirow{2}{*}{Method} & \multicolumn{5}{c}{Efficiency} & \multicolumn{2}{c}{Accuracy (\%)} \\
\cmidrule(lr){3-7} \cmidrule(lr){8-9}
& & \multicolumn{2}{c}{Latency(s)↓} & \multicolumn{2}{c}{TPS↑} & Gen. Length & Flexible↑ & Strict↑ \\
\midrule
\multirow{4}{*}{\begin{tabular}[c]{@{}l@{}}GSM8K \\ \textit{4-shot}\end{tabular}} 
& Vanilla        & 27.48 & 1.00× &  8.44 & 1.00× & 232 / 256 & 78.39 & 37.38 \\ 
& \cellcolor{orange!8}+\textbf{\pad{}}          & \cellcolor{orange!8}18.35 & \cellcolor{orange!8}1.50× & \cellcolor{orange!8}8.76 & \cellcolor{orange!8}1.04× & \cellcolor{orange!8}161 / 256 & \cellcolor{orange!8}78.54 & \cellcolor{orange!8}63.84 \\ 
& +Parallel (Fast-dLLM)      & 8.55 & 3.21× & 27.14 & 3.22× & 232 / 256 & 78.54 & 38.67 \\ 
& \cellcolor{orange!8}+Parallel+\textbf{\pad{}} & \cellcolor{orange!8}\textbf{6.64} & \cellcolor{orange!8}\textbf{4.14×} & \cellcolor{orange!8}24.25 & \cellcolor{orange!8}2.87× & \cellcolor{orange!8}161 / 256 & \cellcolor{orange!8}79.76 & \cellcolor{orange!8}64.97 \\ 
\midrule
\multirow{4}{*}{\begin{tabular}[c]{@{}l@{}}MATH \\ \textit{4-shot}\end{tabular}} 
& Vanilla        & 25.40 & 1.00× & 9.79 & 1.00× & 249 / 256 & 33.58 & 8.42 \\ 
& \cellcolor{orange!8}+\textbf{\pad{}}          & \cellcolor{orange!8}21.61 & \cellcolor{orange!8}1.18× & \cellcolor{orange!8}9.75 & \cellcolor{orange!8}1.00× & \cellcolor{orange!8}211 / 256 & \cellcolor{orange!8}33.42 & \cellcolor{orange!8}28.04 \\ 
& +Parallel (Fast-dLLM)      & 9.91 & 2.56× & 25.09 & 2.56× & 249 / 256 & 33.40 & 8.76 \\ 
& \cellcolor{orange!8}+Parallel+\textbf{\pad{}} & \cellcolor{orange!8}\textbf{9.20} & \cellcolor{orange!8}\textbf{2.76×} & \cellcolor{orange!8}22.93 & \cellcolor{orange!8}2.34× & \cellcolor{orange!8}211 / 256 & \cellcolor{orange!8}33.30 & \cellcolor{orange!8}27.98 \\ 
\midrule
\multirow{4}{*}{\begin{tabular}[c]{@{}l@{}}HumanEval \\ \textit{0-shot}\end{tabular}} 
& Vanilla        & 34.67 & 1.00× & 13.64 & 1.00× & 473 / 512& 43.90 & -- \\ 
& \cellcolor{orange!8}+\textbf{\pad{}}          & \cellcolor{orange!8}27.41 & \cellcolor{orange!8}1.26× & \cellcolor{orange!8}15.96 & \cellcolor{orange!8}1.17× & \cellcolor{orange!8}438 / 512& \cellcolor{orange!8}47.56 & \cellcolor{orange!8}-- \\ 
& +Parallel (Fast-dLLM)      & 11.48 & 3.02× & 41.40 & 3.04× & 475 / 512& 43.29 & -- \\ 
& \cellcolor{orange!8}+Parallel+\textbf{\pad{}} & \cellcolor{orange!8}\textbf{9.14} & \cellcolor{orange!8}\textbf{3.79×} & \cellcolor{orange!8}47.86 & \cellcolor{orange!8}3.51× & \cellcolor{orange!8}438 / 512& \cellcolor{orange!8}46.34 & \cellcolor{orange!8}-- \\ 
\midrule
\multirow{4}{*}{\begin{tabular}[c]{@{}l@{}}MBPP \\ \textit{3-shot}\end{tabular}} 
& Vanilla        & 62.11 & 1.00× & 4.82 & 1.00× & 299 / 512& 15.00 & -- \\ 
& \cellcolor{orange!8}+\textbf{\pad{}}          & \cellcolor{orange!8}15.89 & \cellcolor{orange!8}3.91× & \cellcolor{orange!8}6.85 & \cellcolor{orange!8}1.42× & \cellcolor{orange!8}109 / 512& \cellcolor{orange!8}40.40 & \cellcolor{orange!8}-- \\ 
& +Parallel (Fast-dLLM)      & 14.26 & 4.36× & 20.99 & 4.36× & 299 / 512 & 15.00 & -- \\ 
& \cellcolor{orange!8}+Parallel+\textbf{\pad{}} & \cellcolor{orange!8}\textbf{6.02} & \cellcolor{orange!8}\textbf{10.32×} & \cellcolor{orange!8}18.28 & \cellcolor{orange!8}3.79× & \cellcolor{orange!8}110 / 512 & \cellcolor{orange!8}39.40 & \cellcolor{orange!8}-- \\ 
\midrule

\end{tabular}
\end{table}

\begin{table}[h!]
\footnotesize 
\setlength{\tabcolsep}{4pt} 
\centering
\caption{Performance of LLaDA-1.5 with \pad{} on four benchmarks.}
\label{tb:1.5}
\begin{tabular}{@{}l l rrrr r rr@{}}
\toprule
\multicolumn{9}{c}{LLaDA-1.5} \\
\midrule
\multirow{2}{*}{Benchmark} & \multirow{2}{*}{Method} & \multicolumn{5}{c}{Efficiency} & \multicolumn{2}{c}{Accuracy (\%)} \\
\cmidrule(lr){3-7} \cmidrule(lr){8-9}
& & \multicolumn{2}{c}{Latency(s)↓} & \multicolumn{2}{c}{TPS↑} & Gen. Length & Flexible↑ & Strict↑ \\
\midrule
\multirow{4}{*}{\begin{tabular}[c]{@{}l@{}}GSM8K \\ \textit{4-shot}\end{tabular}} 
& Vanilla        & 27.61 & 1.00× & 7.77 & 1.00× & 215 / 256& 80.59 & 61.87 \\ 
& \cellcolor{orange!8}+\textbf{\pad{}}          & \cellcolor{orange!8}18.26 & \cellcolor{orange!8}1.51× & \cellcolor{orange!8}8.56 & \cellcolor{orange!8}1.10× & \cellcolor{orange!8}156 / 256& \cellcolor{orange!8}80.14 & \cellcolor{orange!8}78.47 \\ 
& +Parallel (Fast-dLLM)      & 8.06 & 3.42× & 26.61 & 3.43× & 215 / 256 & 80.82 & 62.62 \\ 
& \cellcolor{orange!8}+Parallel+\textbf{\pad{}} & \cellcolor{orange!8}\textbf{6.23} & \cellcolor{orange!8}\textbf{4.43×} & \cellcolor{orange!8}25.23 & \cellcolor{orange!8}3.25× & \cellcolor{orange!8}157 / 256& \cellcolor{orange!8}80.89 & \cellcolor{orange!8}78.92 \\ 
\midrule
\multirow{4}{*}{\begin{tabular}[c]{@{}l@{}}MATH \\ \textit{4-shot}\end{tabular}} 
& Vanilla        & 25.12 & 1.00× & 8.67 & 1.00× & 218 / 256& 33.52 & 32.72 \\ 
& \cellcolor{orange!8}+\textbf{\pad{}}          & \cellcolor{orange!8}20.63 & \cellcolor{orange!8}1.22× & \cellcolor{orange!8}9.48 & \cellcolor{orange!8}1.09× & \cellcolor{orange!8}196 / 256& \cellcolor{orange!8}34.08 & \cellcolor{orange!8}37.00 \\ 
& +Parallel (Fast-dLLM)      & 9.48 & 2.65× & 22.96 & 2.65× & 218 / 256 & 33.60 & 32.92 \\ 
& \cellcolor{orange!8}+Parallel+\textbf{\pad{}} & \cellcolor{orange!8}\textbf{8.57} & \cellcolor{orange!8}\textbf{2.93×} & \cellcolor{orange!8}22.76 & \cellcolor{orange!8}2.63× & \cellcolor{orange!8}195 / 512& \cellcolor{orange!8}32.92 & \cellcolor{orange!8}35.96 \\ 
\midrule
\multirow{4}{*}{\begin{tabular}[c]{@{}l@{}}HumanEval \\ \textit{0-shot}\end{tabular}} 
& Vanilla        & 34.80 & 1.00× & 3.16 & 1.00× & 110 / 512& 40.85 & -- \\ 
& \cellcolor{orange!8}+\textbf{\pad{}}          & \cellcolor{orange!8}11.55 & \cellcolor{orange!8}3.01× & \cellcolor{orange!8}7.19 & \cellcolor{orange!8}2.28× & \cellcolor{orange!8}83 / 512& \cellcolor{orange!8}44.51 & \cellcolor{orange!8}-- \\ 
& +Parallel (Fast-dLLM)      & 11.16 & 3.12× & 9.80 & 3.10× & 109 / 512& 39.63 & -- \\ 
& \cellcolor{orange!8}+Parallel+\textbf{\pad{}} & \cellcolor{orange!8}\textbf{5.26} & \cellcolor{orange!8}\textbf{6.61×} & \cellcolor{orange!8}15.64 & \cellcolor{orange!8}4.95× & \cellcolor{orange!8}82 / 512& \cellcolor{orange!8}39.63 & \cellcolor{orange!8}-- \\ 
\midrule
\multirow{4}{*}{\begin{tabular}[c]{@{}l@{}}MBPP \\ \textit{3-shot}\end{tabular}} 
& Vanilla        & 62.34 & 1.00× & 1.02 & 1.00× & 63 / 512& 38.20 & -- \\ 
& \cellcolor{orange!8}+\textbf{\pad{}}          & \cellcolor{orange!8}14.95 & \cellcolor{orange!8}4.17× & \cellcolor{orange!8}4.33 & \cellcolor{orange!8}4.26× & \cellcolor{orange!8}65 / 512& \cellcolor{orange!8}39.80 & \cellcolor{orange!8}-- \\ 
& +Parallel (Fast-dLLM)      & 5.47 & 11.39× & 11.62 & 11.44× & 64 / 512& 38.60 & -- \\ 
& \cellcolor{orange!8}+Parallel+\textbf{\pad{}} & \cellcolor{orange!8}\textbf{4.41} & \cellcolor{orange!8}\textbf{14.14×} & \cellcolor{orange!8}14.83 & \cellcolor{orange!8}14.60× & \cellcolor{orange!8}65 / 512& \cellcolor{orange!8}41.60 & \cellcolor{orange!8}-- \\ 
\midrule
\end{tabular}
\end{table}

\begin{table}[h!]
\footnotesize 
\setlength{\tabcolsep}{4pt} 
\centering
\caption{Performance of Dream-Base with \pad{} on four benchmarks.}
\label{tb:dream}
\begin{tabular}{@{}l l rrrr r rr@{}}
\toprule
\multicolumn{9}{c}{Dream-Base} \\
\midrule
\multirow{2}{*}{Benchmark} & \multirow{2}{*}{Method} & \multicolumn{5}{c}{Efficiency} & \multicolumn{2}{c}{Accuracy (\%)} \\
\cmidrule(lr){3-7} \cmidrule(lr){8-9}
& & \multicolumn{2}{c}{Latency(s)↓} & \multicolumn{2}{c}{TPS↑} & Gen. Length & Flexible↑ & Strict↑ \\
\midrule
\multirow{4}{*}{\begin{tabular}[c]{@{}l@{}}GSM8K \\ \textit{4-shot}\end{tabular}} 
& Vanilla        & 22.30 & 1.00× & 11.43 & 1.00× & 255 / 256& 75.06 & 74.37 \\ 
& \cellcolor{orange!8}+\textbf{\pad{}}          & \cellcolor{orange!8}10.27 & \cellcolor{orange!8}2.17× & \cellcolor{orange!8}12.75 & \cellcolor{orange!8}1.11× & \cellcolor{orange!8}131 / 256& \cellcolor{orange!8}75.28 & \cellcolor{orange!8}75.06 \\ 
& +Parallel (Fast-dLLM)      & 13.84 & 1.61× & 18.43 & 1.61× & 255 / 256 & 75.51 & 74.83 \\ 
& \cellcolor{orange!8}+Parallel+\textbf{\pad{}} & \cellcolor{orange!8}\textbf{5.24} & \cellcolor{orange!8}\textbf{4.25×} & \cellcolor{orange!8}24.17 & \cellcolor{orange!8}2.11× & \cellcolor{orange!8}127 / 256& \cellcolor{orange!8}74.83 & \cellcolor{orange!8}74.75 \\ 
\midrule
\multirow{4}{*}{\begin{tabular}[c]{@{}l@{}}MATH \\ \textit{4-shot}\end{tabular}} 
& Vanilla        & 21.01 & 1.00× & 12.19 & 1.00× & 256 / 256& 34.06 & 37.76 \\ 
& \cellcolor{orange!8}+\textbf{\pad{}}          & \cellcolor{orange!8}16.64 & \cellcolor{orange!8}1.26× & \cellcolor{orange!8}15.33 & \cellcolor{orange!8}1.26× & \cellcolor{orange!8}255 / 256& \cellcolor{orange!8}34.14 & \cellcolor{orange!8}37.64 \\ 
& +Parallel (Fast-dLLM)      & 8.82 & 2.38× & 29.03 & 2.38× & 256 / 256 & 35.12 & 38.62 \\ 
& \cellcolor{orange!8}+Parallel+\textbf{\pad{}} & \cellcolor{orange!8}\textbf{7.72} & \cellcolor{orange!8}\textbf{2.72×} & \cellcolor{orange!8}33.04 & \cellcolor{orange!8}2.71× & \cellcolor{orange!8}255 / 256& \cellcolor{orange!8}34.44 & \cellcolor{orange!8}38.32 \\ 
\midrule
\multirow{4}{*}{\begin{tabular}[c]{@{}l@{}}HumanEval \\ \textit{0-shot}\end{tabular}} 
& Vanilla        & 28.49 & 1.00× & 17.93 & 1.00× & 511 / 512& 51.22 & -- \\ 
& \cellcolor{orange!8}+\textbf{\pad{}}          & \cellcolor{orange!8}8.20 & \cellcolor{orange!8}3.47× & \cellcolor{orange!8}26.83 & \cellcolor{orange!8}1.50× & \cellcolor{orange!8}220 / 512 & \cellcolor{orange!8}51.22 & \cellcolor{orange!8}-- \\ 
& +Parallel (Fast-dLLM)      & 14.15 & 2.01× & 36.11 & 2.01× & 511 / 512& 53.05 & -- \\ 
& \cellcolor{orange!8}+Parallel+\textbf{\pad{}} & \cellcolor{orange!8}\textbf{4.06} & \cellcolor{orange!8}\textbf{7.01×} & \cellcolor{orange!8}52.62 & \cellcolor{orange!8}2.93× & \cellcolor{orange!8}214 / 512& \cellcolor{orange!8}52.44 & \cellcolor{orange!8}-- \\ 
\midrule
\multirow{4}{*}{\begin{tabular}[c]{@{}l@{}}MBPP \\ \textit{3-shot}\end{tabular}} 
& Vanilla        & 49.15 & 1.00× & 10.42 & 1.00× & 512 / 512 & 52.40 & -- \\ 
& \cellcolor{orange!8}+\textbf{\pad{}}          & \cellcolor{orange!8}41.36 & \cellcolor{orange!8}1.19× & \cellcolor{orange!8}12.38 & \cellcolor{orange!8}1.19× & \cellcolor{orange!8}512 / 512& \cellcolor{orange!8}52.60 & \cellcolor{orange!8}-- \\ 
& +Parallel (Fast-dLLM)      & 12.38 & 3.97× & 41.36 & 3.97× & 512 / 512& 55.40 & -- \\ 
& \cellcolor{orange!8}+Parallel+\textbf{\pad{}} & \cellcolor{orange!8}\textbf{9.86} & \cellcolor{orange!8}\textbf{4.98×} & \cellcolor{orange!8}51.92 & \cellcolor{orange!8}4.98× & \cellcolor{orange!8}512 / 512& \cellcolor{orange!8}54.80 & \cellcolor{orange!8}-- \\ 
\bottomrule
\end{tabular}
\end{table}

\paragraph{Latency.}  
The latency improvements of \pad{} arise from three sources.  
(1) {Reduced suffix complexity:} in standard dLLMs, handling suffix tokens incurs quadratic complexity with respect to suffix length, whereas \pad{} reduces this to linear complexity via suffix dropout.  
(2) {Early termination:} by decoupling generation from a fixed sequence length through our sliding window design, \pad{} naturally supports early stopping once the end-of-sequence is reached.  
(3) {More concise generations:} suffix dropout removes low-entropy tokens, leading the model to produce more compact outputs and slightly shorter generations. Importantly, as analyzed in the Accuracy section, this does not degrade task accuracy.  
Across most benchmarks, the reduction in generation length is modest (e.g., on LLaDA-1.5, about $10\%$ on MATH and $27\%$ on GSM8K), while datasets such as MBPP remain almost unchanged.  
Thus, the majority of latency gains come from reduced suffix computation rather than shorter outputs.  

Overall, \pad{} consistently reduces latency across all benchmarks in the three-model dLLM suite, achieving $\mathbf{1.18\times}$ to $\mathbf{4.17\times}$ speedups over the vanilla backbone.  
When combined with parallel decoding, \pad{} yields an additional $1.08\times$ to $3.48\times$ speedup over parallel decoding alone, resulting in an overall speedup of $\mathbf{2.76\times}$ to $\mathbf{14.14\times}$ over vanilla.  
This complementarity arises because the two methods target orthogonal bottlenecks: \pad{} eliminates redundant KV-token computation via suffix dropout, while parallel decoding in Fast-dLLM~\cite{fast-dllm} mitigates dependency constraints by selectively decoding only high-confidence tokens in parallel.  
By combining these approaches, we exploit both finer-grained token pruning and safe multi-token prediction, yielding substantial efficiency gains.  

\begin{figure}[t]
\hspace{-6mm}
    \centering
    \includegraphics[width=1\linewidth]{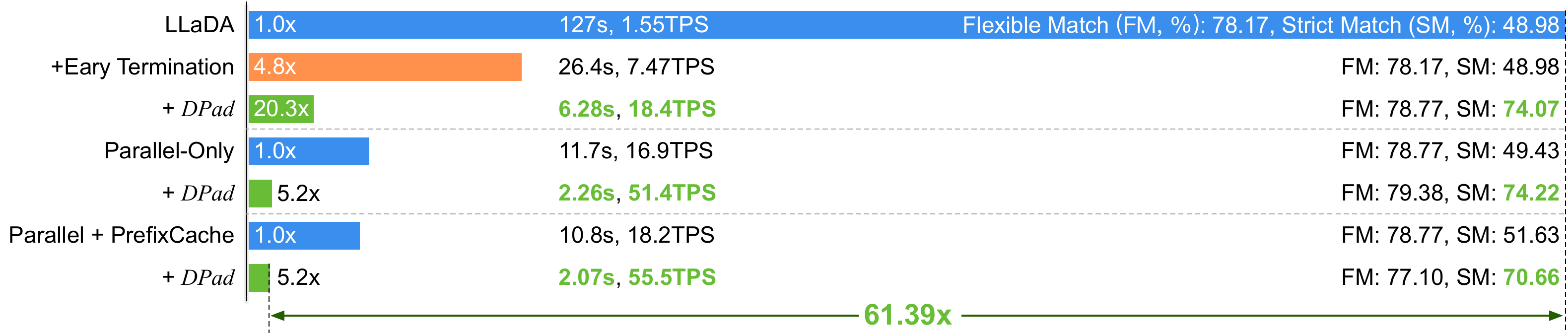}
    \caption{Latency comparison on {LLaDA-1.5} with GSM8K (1024 tokens, 1-shot).}
    \label{fig:speedup_llada}
\end{figure}

That said, the advantage of suffix dropout is less pronounced on short-sequence benchmarks.  
This is due to two factors.  
First, our Gaussian-based dropout typically yields about $62.5\%$ sparsity within the sliding window, so the constant overhead remains significant.  
It is also impractical to shrink the window much further when the maximum generation length itself is short (e.g., 256).  
Second, when the prompt dominates the sequence (e.g., $\sim$80\% prompt vs. 20\% suffix in GSM8K and MATH), suffix attention accounts for only a small fraction of total computation.  
According to Amdahl’s law~\cite{amdahl1967validity}, the maximum achievable speedup in such cases is inherently bounded (often around $1.1\times$).  
Nevertheless, even without any system-level optimizations, our simple implementation of suffix dropout consistently delivers stable latency improvements.  
Its true potential emerges in \textbf{longer-sequence} settings in \Fig{fig:speedup_llada} and \Fig{fig:speedup_dream}, where the suffix fraction grows; as we show later with 1024-token sequences, \pad{} achieves substantial additional reductions in latency.

\begin{figure}[t]
    \centering
    \includegraphics[width=1\linewidth]{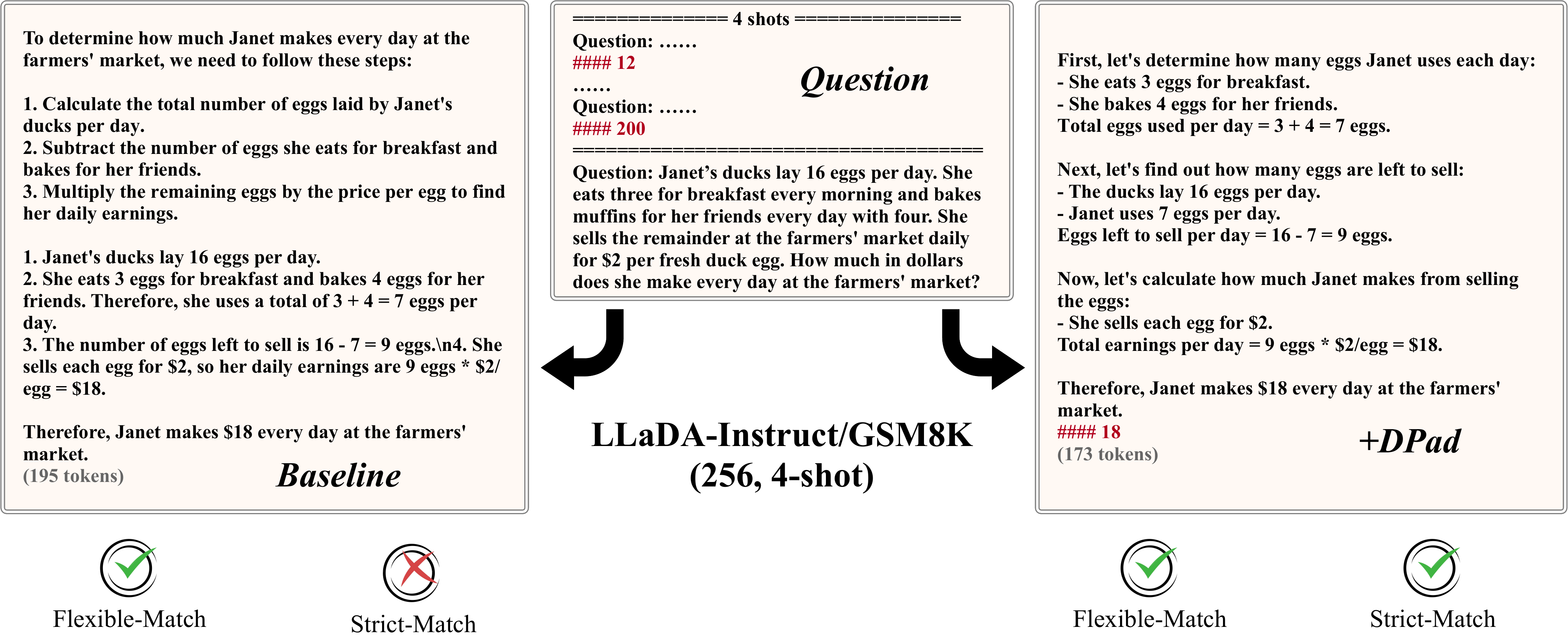}
   \caption{{A Case Study from GSM8K on In-Context Learning and Format Adherence.} The figure contrasts a baseline model's output with the same model enhanced by \pad{}. The baseline produces the correct answer (passing Flexible-Match) but fails to replicate the structured reasoning from the prompt, thus failing the Strict-Match. \pad{} successfully generates both the correct answer and the required format, passing both evaluations.}
    \label{fig:strict}
\end{figure}

\paragraph{Throughput.}  
We observe a subtle distinction between latency- and TPS-based efficiency metrics.  
While \pad{} consistently reduces per-sample latency, its TPS gains may appear less pronounced.  
This is because \pad{} often encourages the model to generate more concise and complete responses, reaching the end-of-sequence earlier and producing fewer redundant tokens.  
As illustrated in \Fig{fig:strict} and \Fig{fig:case2}, this behavior reflects not a limitation but an improvement: the model terminates naturally rather than exhausting its context with low-quality continuations.  
We view this as an additional quality benefit of \pad{}, complementing its efficiency gains.  

However, when the generation length is reduced more substantially, TPS can drop.  
This mainly occurs because we apply no additional system- or algorithm-level optimizations: shorter generations lead to lower GPU utilization, which mechanically reduces TPS.  
Yet, the extra tokens produced by baselines often carry low semantic information; higher TPS simply reflects more tokens, not greater efficiency.  
For example, if we were to artificially pad outputs with meaningless characters, TPS would increase, but this would not represent a genuine efficiency gain.  
Therefore, in our main results, we report raw throughput numbers without post-processing.  
This also highlights a broader point: the community may need to reconsider throughput metrics and develop alternatives that better balance sequence length and accuracy, rewarding models that achieve comparable accuracy with shorter generations.

\paragraph{Accuracy.}
In addition to improving inference efficiency, \pad{} also enhances accuracy across nearly all tasks for the LLaDA-Instruct and LLaDA-1.5 models (Table 1), thereby defying the typical trade-off between speed and accuracy. For instance, \pad{} yields substantial gains in strict-match accuracy on GSM8K (+26.46\%) and MATH (+19.62\%) for LLaDA-Instruct.
By contrast, the Dream-Base model shows no consistent advantage, with accuracy broadly comparable to the baseline and fluctuating only within a narrow margin.  
We attribute this stability to differences in training protocols, particularly the absence of instruction tuning such as supervised fine-tuning (SFT).

The improvement in strict-match score is particularly noteworthy, as it highlights \pad{}’s ability to enhance in-context learning. The vanilla backbone typically exhibits low strict-match performance (e.g., only 37.38\% on GSM8K for LLaDA-Instruct), since this metric requires the model not only to produce the correct final answer (Flexible-Match) but also to adhere to the specific reasoning format demonstrated in few-shot exemplars, as illustrated in Figure~\ref{fig:strict}. We posit that failures in strict matching often stem from interference by distant suffix tokens, which introduce low-value or off-format patterns that distract the model and encourage verbose or poorly structured outputs. By reducing the influence of such suffix tokens and directing attention toward high-value, information-rich prefix exemplars, \pad{} enables the model to replicate the structured reasoning formats required by strict matching more faithfully. In addition, by suppressing redundant generations, \pad{} facilitates earlier convergence to concise, well-formatted outputs, further improving strict-match performance without altering model parameters.

\subsection{Ablations and Analysis}  
\subsubsection{Maximum Generation Length}
\label{sec:speedup}

\begin{figure}[t]
\hspace{-6mm}
    \centering
    \includegraphics[width=1\linewidth]{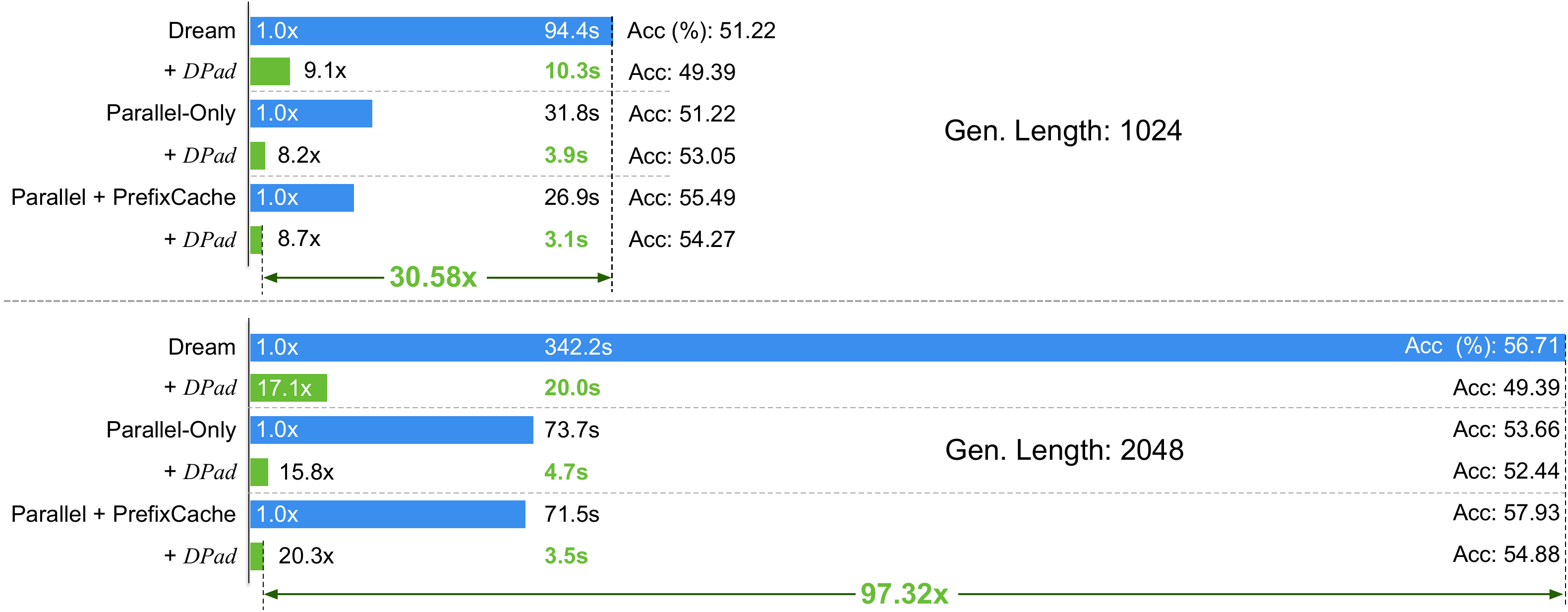}
    \caption{Latency comparison on {Dream-Base} with HumanEval (1024 and 2048 tokens, 1-shot).}
    \label{fig:speedup_dream}
\end{figure}

To better quantify the efficiency gains of different acceleration strategies in long-sequence generation,  
we analyze speedups under various configurations, as shown in \Fig{fig:speedup_llada} and \Fig{fig:speedup_dream}.  
Specifically, we evaluate GSM8K with LLaDA-1.5 and HumanEval with Dream under the following strategies:  
\texttt{Vanilla}, \texttt{Vanilla+ET} (augmented with an early-termination mechanism), \texttt{+Pad} (our method alone), and \texttt{+Parallel} (Fast-dLLM).  
We then consider combined strategies to assess complementarity, including \texttt{+Pad+Parallel}, \texttt{+Parallel+PrefixCache}, and \texttt{+Parallel+PrefixCache+Pad}.  
This setup allows us to disentangle the contributions of suffix dropout, parallel decoding, and prefix caching, as well as their interactions.

\paragraph{Speedup.}  
We find that the acceleration benefits of \pad{} grow substantially with sequence length.  
For LLaDA-1.5 on GSM8K, improvements are modest at shorter sequence lengths (up to $1.51\times$ under a 256-token limit).  
However, when the maximum length is extended to 1024 tokens (single-shot setting), standalone \pad{} achieves a dramatic $\mathbf{20.3\times}$ speedup.  
This effect arises because vanilla LLaDA typically produces concise answers (about 200 tokens) but continues generating redundant \texttt{<eos>} tokens to fill the context window, incurring wasted computation.  
To provide a stronger baseline, we augment vanilla with early termination; even under this setting, \pad{} still delivers a $\mathbf{4.8\times}$ improvement.  
Finally, when combined with parallel decoding and prefix caching (Fast-dLLM), the efficiency gains compound, yielding an overall $\mathbf{61.39\times}$ speedup compared to vanilla LLaDA and a $\mathbf{8.7\times}$ improvement over Fast-dLLM.

\begin{figure}[t]
    \centering
    \includegraphics[width=0.9\linewidth]{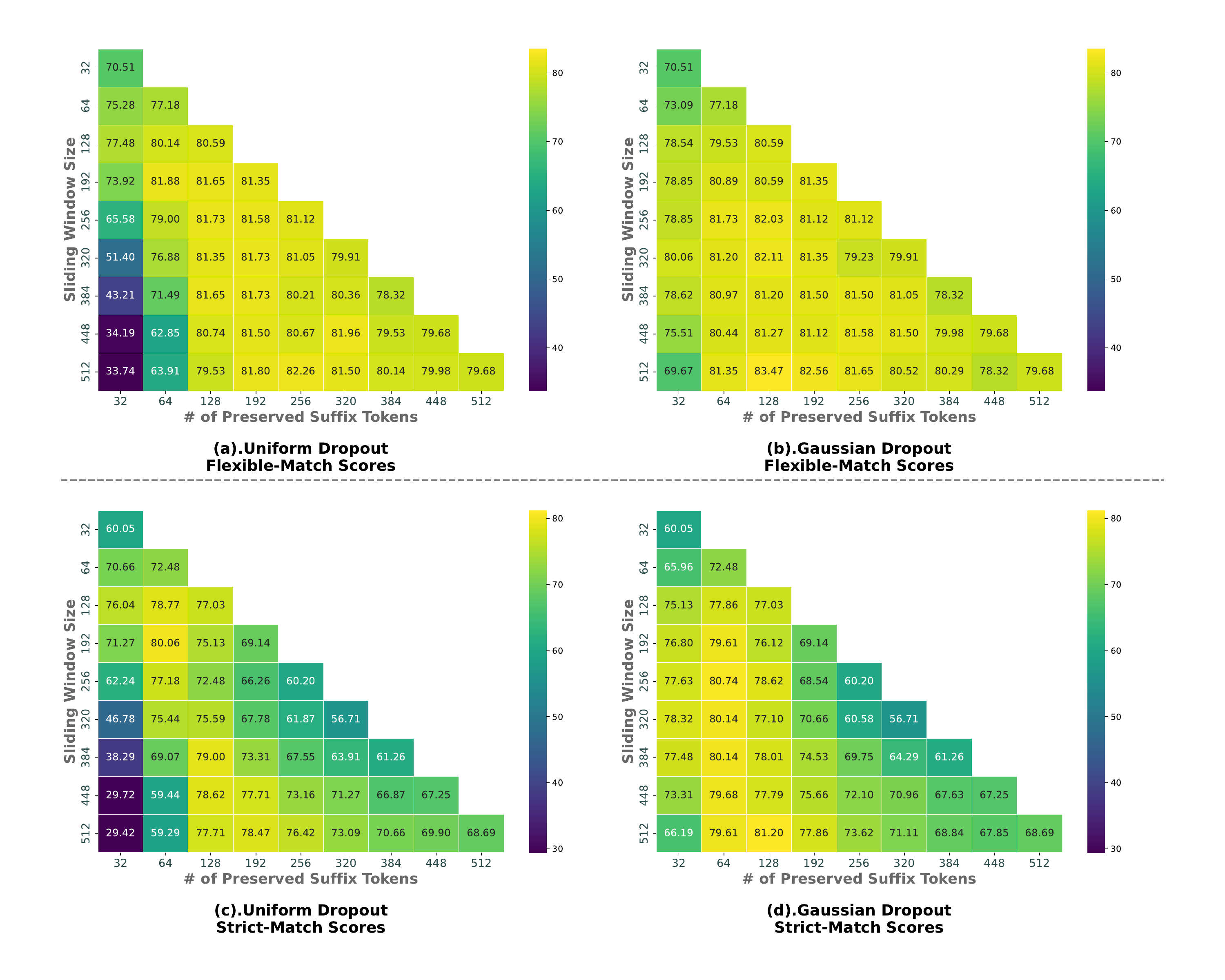}
    \caption{{Ablation Study on Sliding Window Size and Dropout Function for \pad{} on LLaDA-1.5/GSM8K (512, 4-shot). }Heatmaps showing Flexible-Match Accuracy scores with (a) uniform and (b) Gaussian dropout, and Strict-Match Accuracy scores with (c) uniform and (d) Gaussian dropout, across varying sliding window sizes and number of preserved suffix tokens. The (512, 512) configuration corresponds to the baseline, as it involves no token dropout.}
    \label{fig:window_function}
\end{figure}

A similar scaling trend is observed with Dream-Base on HumanEval.  
Here, \pad{} alone accelerates inference by $\mathbf{9.13\times}$ for 1024-token sequences and $\mathbf{17.1\times}$ for 2048-token sequences.  
When combined with Fast-dLLM, the benefits become multiplicative: $\mathbf{30.58\times}$ at 1024 tokens and $\mathbf{97.32\times}$ at 2048 tokens.  
These results demonstrate that suffix dropout and parallel decoding address orthogonal bottlenecks and, when combined, yield near two orders of magnitude improvement in long-sequence generation.  
Overall, the strong scaling with sequence length highlights \pad{} as a key component for enabling dLLMs to match the scalability of traditional autoregressive models.

\paragraph{Accuracy.}
Longer, low-shot settings further highlight \pad{}'s ability to preserve and even enhance model accuracy. In the 1-shot LLaDA setting, the strict-match score of the baseline drops significantly (from $\sim$60\% to $\sim$50\%), whereas \pad{}'s performance remains remarkably stable (dropping only from $\sim$78\% to $\sim$74\%). This resilience demonstrates that \pad{} substantially strengthens the model's in-context learning capability, a significant achievement for a training-free method.

By contrast, on the 2048-token HumanEval task, we observed a 7.32\% accuracy degradation when applying \pad{} to the vanilla Dream model. This degradation is largely mitigated when \pad{} is combined with Fast-dLLM. We hypothesize that this isolated performance drop arises from complex interactions between our training-free pruning strategy and the model's native Top-$k$ sampling behavior, pointing to an interesting direction for future investigation.

\subsubsection{Sliding Window Size and Dropout Function}
\label{sec:window_function}

We conducted an ablation study on LLaDA-1.5/GSM8K to determine the optimal sliding window size and dropout function for \pad{}. As shown in Figure~\ref{fig:window_function}, we evaluated both flexible-match and strict-match scores while varying the window size, the number of preserved suffix tokens, and the dropout function (Uniform vs. Gaussian).

Our analysis reveals two key findings that point to the existence of a \textbf{critical context window} of approximately 64--128 tokens immediately following the current block. First, within this critical window, performance consistently improves as more suffix tokens are preserved. For windows extending beyond this zone, however, performance exhibits a concave trend, peaking at a density of around 50\%. Second, when the token budget is limited (e.g., fewer than 128 preserved tokens), spreading this budget thinly across a larger window significantly degrades accuracy.

These observations lead to a clear principle for our method: the token budget should be prioritized to maintain a high density of preserved tokens within the critical 64--128 token window, as expanding the window to more distant tokens with a limited budget can be counterproductive.

As further evidence, \Fig{fig:window_function} validates our choice of a truncated Gaussian function for token dropout. Compared against a uniform random baseline, Gaussian dropout performs comparably under large token budgets ($\geq192$), but its advantage becomes increasingly clear under more stringent, low-budget conditions. In particular, when the sliding window is large, \textbf{Gaussian dropout consistently outperforms the uniform strategy}, achieving superior or equivalent performance with fewer preserved tokens.

This empirical result aligns with the consistent decaying patterns observed in RoPE~\cite{su2021roformer}, attention scores (\Fig{fig:attention_score}), and token confidence maps~\cite{gong_diffucoder_2025}, which collectively motivate our design.  
By biasing dropout toward nearby tokens, the Gaussian approach allocates the token budget more efficiently and thus delivers better performance under limited computation.  
Of course, Gaussian sampling may not be the optimal decay function, and other decay-based schemes (e.g., exponential, linear, or step-wise cutoff) remain to be explored.  
Nevertheless, in the training-free setting, we find that results are largely insensitive to the exact decay form, as long as the scheme emphasizes nearer tokens.

\subsubsection{The Choice of Gaussian Hyperparameters}
Inspired by findings that dLLMs exhibit different behaviors on mathematical and code-generation tasks~\cite{gong_diffucoder_2025}, we tune the hyperparameters for our Gaussian Sampler separately for each domain. We perform a grid search over two key parameters: the decay rate, $k$, and a scale factor, $a$. The parameter $a$ is used to control the overall retention {density}, which we define as the expected proportion of suffix tokens preserved by the Gaussian Sampler inside the sliding window.

\begin{figure}[h]
    \centering
    \includegraphics[width=1\linewidth]{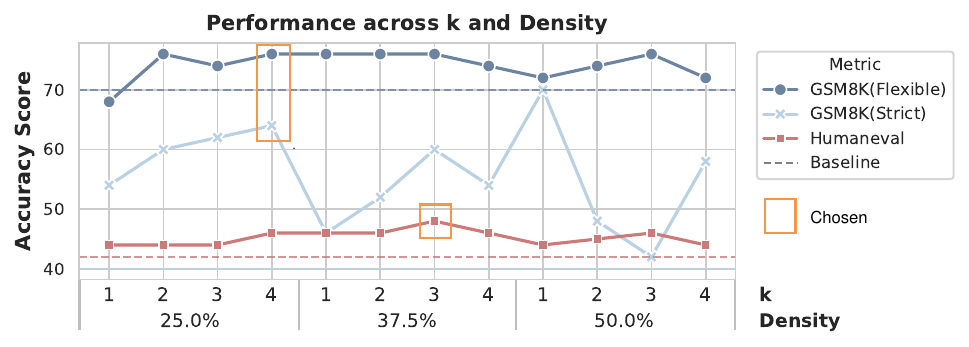}
    \caption{{Ablation study on hyperparameters $k$ and $a$ for LLaDA-Instruct on subsets of GSM8K and HumanEval.} The parameter $a$ is mapped to the retention density shown on the x-axis. Each dashed line represents the baseline performance for the solid line of the same color and metric.}
    \label{fig:ka}
\end{figure}

The results of this search for LLaDA-Instruct are presented in Figure~\ref{fig:ka}, conducted on 50-sample subsets of GSM8K and HumanEval. While using a small subset for tuning may introduce some variance, the findings provide clear directional insights. Our method with Gaussian Dropout consistently outperforms the baseline in nearly all configurations, with a single, minor exception for the GSM8K flexible-match accuracy score at $k=1$ and a density of 25.0\%. The suboptimal performance at $k=1$ is expected, as at this value the Gaussian curve is relatively flat across the sampling window (see Appendix~\ref{appx:hyper}), causing the sampling to degenerate into a near-uniform distribution.

Focusing on the more effective range of $k \in [2,4]$, we identify distinct optimal settings for each domain. For the mathematical reasoning task (GSM8K), a density of 25.0\% provides the best balance of accuracy and efficiency. Although some settings with 50.0\% density achieve a slightly higher strict-match score, they do so by preserving twice as many suffix tokens, which significantly undermines acceleration. We therefore select $k=4.0$ and a density of 25.0\% (from $a=2.0$) as the optimal configuration. For code generation (HumanEval), a configuration of $k=3.0$ and a density of 37.5\% (from $a=2.3$) yields the best performance.

Based on these findings, we adopt these hyperparameters for all subsequent math and code benchmarks for LLaDA-Instruct (see Appendix~\ref{appx:hyper}), with similar tuning for other models.

\section{Discussion}

\subsection{Beyond a Training-free Method: Suffix Dropout with SFT}

While \pad{} demonstrates improved accuracy and computational efficiency across diverse benchmarks by strategically pruning the suffix space, we observe a performance degradation in very long-sequence generation, particularly at a context length of 2048 tokens.  
We attribute this behavior to a distributional shift introduced by suffix dropout.  
During pre-training, the model is optimized to predict token distributions conditioned on a full, continuous sequence of masked suffix tokens.  
In contrast, our distance-decay dropout forces the model at inference time to predict based on a much smaller, discontinuous set of masked suffix tokens.  
While moderate dropout helps the model focus on prompt semantics, such a large shift in the conditional distribution can hinder generalization, leading to degraded performance.  

This training–inference distribution gap may be mitigated with supervised finetuning (SFT).  
In particular, the attention lottery hypothesis offers a useful perspective that can be incorporated into training.  
We revise the learning objective as:
\begin{equation}
\mathcal{L}_{\text{\pad{}}}(\theta) = -\mathbb{E}_{{x_0}, t, M}\left[\sum_{i \in C} \log p_{\theta}\left({x_0}^{i} \mid {x_t}^{(I_R \cup I_P \cup M)}\right)\right],
\end{equation}
where $M$ denotes a subset of masked suffix tokens sampled via distance-decay dropout at each training step, $I_R$ represents the prompt tokens, $I_p$ represents the generated prefix tokens,$\theta$ represents the model parameters, $t \in [0,1]$ is degree of masking of samples from the forward masking process (\Sec{dLLMfoundamentals}), and the loss is computed over the current block tokens $C$.  
Using stochastic dropout masks rather than a fixed deterministic pattern enhances robustness, as the model does not overfit to any single dropout scheme.  
This revised objective explicitly integrates the notion of an attention lottery into training, encouraging the model to avoid wasting capacity by writing redundant information into distant suffix tokens that are likely to be pruned at inference.  

Looking further ahead, one could even incorporate distance-decay dropout directly into the pre-training phase, allowing the model to learn sparsity from scratch.  
Such pre-training with sparse suffix attention would naturally align training and inference conditions, and may yield even stronger efficiency–accuracy trade-offs.

\subsection{Comparison to Semi-Autoregressive Diffusion and Block Diffusion}

Block Diffusion models~\cite{arriola2025blockdiffusioninterpolatingautoregressive} operate autoregressively at the block level, predicting each block conditioned only on its predecessors.  
As a result, their attention mechanism excludes access to subsequent blocks, in contrast to the directional attention employed by semi-autoregressive models.  

While this design is computationally efficient, its strictly forward-looking nature introduces two major limitations.  
First, it is vulnerable to the reversal curse~\cite{berglund2024reversalcursellmstrained}, since it cannot capture long-range (beyond a single block) bidirectional dependencies that are crucial for tasks such as code generation, which often require iterative back-and-forth refinement~\cite{gong_diffucoder_2025}.  
Second, by discarding the suffix, it forfeits the ability to use suffix tokens as a \emph{scratchpad} (see \Sec{scratchpad}), thereby losing an important medium for contextual organization.  

Our proposed distance-decay dropout interpolates between these paradigms.  
It retains suffix tokens to function as a scratchpad, while at the same time preserving the model’s bidirectional learning ability, allowing information to be both written into and retrieved from the suffix scratchpad as needed.
\section{Conclusion}
We addressed a key bottleneck in dLLMs: the high cost of full suffix attention, where most distant tokens are redundant and add little value.  
To overcome this, we introduced the Diffusion Scratchpad (\pad{}), a simple training-free inference strategy that redefines suffix attention.  
\pad{} combines a fixed-length sliding window, reducing complexity from quadratic to linear, with a distance-decay dropout that prunes low-entropy suffix tokens before computation.  
This design leverages the inherent sparsity of suffix attention, constructing an efficient “winning ticket” for generation on the fly. 
Experiments show that \pad{} consistently improves efficiency without sacrificing accuracy, achieving up to $61.4\times$ speedups when combined with existing optimizations.  
These results highlight \pad{} as a practical and scalable solution for long-sequence generation, paving the way for the dLLM to move from a promising alternative to a viable foundation for future applications.

\bibliographystyle{plain}
\bibliography{references.bib}

\appendix
\section{Appendix}

\subsection{Experiment Details}
\label{appx:hyper}

This section reports the hyperparameters used for our Gaussian sampler, namely $k$ and $a$, across all experiments.  
We observe that attention score distributions vary across datasets, even for the same model.  
As a post-training method, it is difficult for \pad{} to adopt a single set of hyperparameters that works universally across models and benchmarks.  
Therefore, for each benchmark, we sample a small subset of data to determine the hyperparameters that perform best for the corresponding model.  
The hyperparameters used in \Sec{sec:main_results} are summarized in \Tbl{tb:hyper}.  

\begin{table}[h!]
\centering
\small 
\setlength{\tabcolsep}{5pt} 
\renewcommand{\arraystretch}{1.4} 
\caption{The hyperparameters for Gaussian Sampler used in main experiments in \Sec{sec:main_results}}
\begin{tabular}{lllllllllllll}
\toprule
\multirow{2}{*}{Task} & \multicolumn{4}{c}{LLaDA-Instruct}                                                        & \multicolumn{4}{c}{LLaDA-1.5}                                                             & \multicolumn{4}{c}{Dream-Base}                                                            \\ \cline{2-13} 
                      & $k$                  & $a$                  & Density              & Window               & $k$                  & $a$                  & Density              & Window               & $k$                  & $a$                  & Density              & Window               \\ \hline
GSM8K    & 4.0 & 2.0 & 25.0\% & 256 & 3.0 & 1.6 & 25.0\% & 256 & 4.0 & 1.6 & 20.0\% & 256 \\ \hline
\rowcolor{orange!8}   Math     & 4.0 & 2.0 & 25.0\% & 256 & 3.0 & 1.6 & 25.0\% & 256 & 4.0 & 1.6 & 20.0\% & 128 \\ \hline
  HumanEval& 3.0 & 2.3 & 37.5\% & 512 & 3.0 & 1.6 & 25.0\% & 128 & 3.0 & 2.3 & 37.5\% & 128 \\ \hline
\rowcolor{orange!8}   MBPP     & 3.0 & 2.3 & 37.5\% & 128 & 3.0 & 1.6 & 25.0\% & 512 & 3.0 & 1.6 & 25.0\% & 128 \\
\bottomrule
\end{tabular}
\label{tb:hyper}
\end{table}

\subsection{Case Studies}

\begin{figure}[h!]
    \centering
    \includegraphics[width=0.8\linewidth]{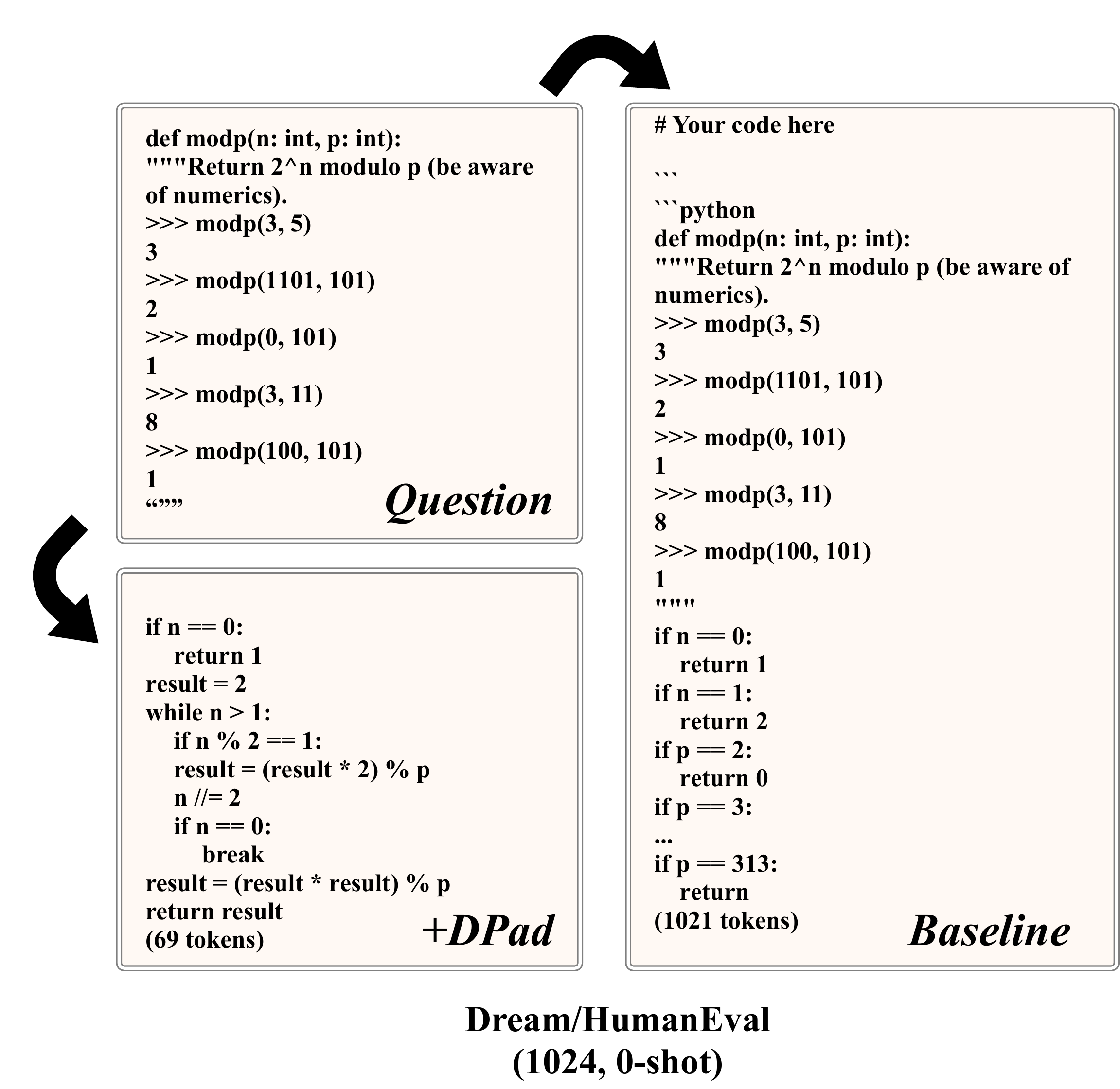}
    \caption{{Dream/HumanEval (1024, 0-shot).}  
    Comparison of generated code for a mathematical problem.  
    The baseline model (1021 tokens) brute-forces the solution by unrolling the loop to fill its token budget.  
    In contrast, \pad{} (69 tokens) generates a concise and logically sound solution that terminates naturally, demonstrating improved reasoning and the avoidance of unnecessary token generation.}
    \label{fig:case2}
\end{figure}

We further present a case study on the Dream model to illustrate how \pad{} influences generation behavior.  
Unlike the LLaDA models, which often terminate responses naturally, the Dream model tends to exhaust its full token budget, even when shorter responses would suffice.  
This results in verbose and computationally inefficient outputs.  
As we demonstrate, \pad{} mitigates this tendency by encouraging the model to generate concise and logically sound solutions.  
Figure~\ref{fig:case2} provides a representative example.  

\end{document}